# Stochastic Simulation of Bayesian Belief Networks

Homer L. Chin and Gregory F. Cooper

Medical Computer Science Group
Knowledge Systems Laboratory
Medical School Office Building, Room 215
Stanford University Medical Center
Stanford, California 94305

## Abstract

This paper examines the use of stochastic simulation of Bayesian belief networks as a method for computing the probabilities of values of variables. Specifically, it examines the use of a scheme described by Henrion, called logic sampling, and an extension to that scheme described by Pearl. The scheme devised by Pearl allows us to "clamp" any number of variables to given values and to conduct stochastic simulation on the resulting network. We have found that this algorithm, in certain networks, leads to much slower than expected convergence to the true posterior probability. This behavior is a result of the tendency for local areas in the graph to become fixed through many stochastic iterations. The length of this non-convergence can be made arbitrarily long by strengthening the dependency between two nodes. This paper describes the use of *graph modification*. By modifying a belief network through the use of pruning, arc reversal, and node reduction, it may be possible to convert the network to a form that is computationally more efficient for stochastic simulation.

## 1. Introduction

The graphical representation of probabilistic relationships between events has been the subject of considerable research. In the field of artificial intelligence, numerous systems have used a directed graph to represent probabilistic relationships between events [Duda 76, Weiss 78]. A particular probabilistic graphical representation has been independently defined and explored by several researchers. In addition to being called belief networks [Pearl 86], they have been termed *causal nets* [Good 61], *probabilistic cause-effect models* [Rousseau 68], *probabilistic causal networks* [Cooper 84], and *influence diagrams* [Howard 84, Shachter 86].

Graphical representation of probabilistic relationships allows for the efficient representation of all the existing dependencies between variables. It is only necessary to consider the known dependencies among the variables in a domain, rather than assuming that all variables are dependent on all other variables. This leads to a significant decrease in the number of probabilities needed to define the outcome space, and improved computational efficiency.

Although probabilistic graphical representations often allow efficient probabilistic inference, some inference problems involving particular topological classes of belief networks have been resistant to any efficient algorithmic solution [Pearl 86]. *Multiply-connected networks*, the most general class of such problems, belong to a class of difficult problems which have are resistant to any general, efficient solutions, and have been shown to belong to the class of NP-hard problems [Cooper 87]. All known exact algorithms for performing inference over multiply-connected belief networks are exponential in the size of the belief network.

## 2. Stochastic Simulation of Bayesian Belief Networks

Because exact solution of multiply connected networks is exponentially complex in the worst cases, the development of stochastic simulation techniques to generate probabilities for variables in the network has been an area of considerable research interest. One such method, presented by Henrion [Henrion 86], is *logic sampling*. In a Bayesian belief network where variables are represented by nodes, simulation is begun with the nodes that have no parent nodes. A value is assigned to each such node based on its prior probability of occurrence. For example, if the probabilities for the values of a node are P(HIGH) = 0.4, P(MEDIUM) = 0.4, and P(LOW) = 0.2, then a value for this node can be simulated by generating a random number between 0 and 1.0. For numbers between 0 and 0.4, the node is set to the value HIGH; for numbers between 0.4 and 0.8, the node is set to MEDIUM; and for numbers above 0.8, the node is set to LOW. The children of these nodes can be similarly assigned values based on the conditional probabilities relating them to their parents. This process is continued until all the nodes in the network have been simulated.

After many simulations, the probability of a value for a given variable can be approximated by dividing the number of times a variable is assigned a given value by the number of simulations performed. The proximity of this approximation to the true probability of a value for a node can be determined by calculating the statistical variance.

## 2.1. Simulation of a Partially Instantiated Network

The difficulty with this technique lies in attempting to perform stochastic simulation in a network where some of the variables have known values. For example, given a belief network that represents medical knowledge, with nodes representing diseases and symptoms, we might want to know the probability of a disease given the occurrence of certain symptoms.

The technique of stochastic simulation described by Henrion does not allow us to fix the value of any nodes other than those that have no parent nodes. For this reason, it does not lend itself well to determining the probability of certain nodes given the occurrence of others. One solution is to perform the simulation multiple times until it produces an instance in which all variables with known values have been assigned those values. This instance is then counted as one positive instance of a simulation, and the process is repeated until enough positive instances are generated to be statistically significant. In those cases where the combination of findings are rare, an inordinate number of simulations may be needed to produce one positive instance.

Pearl cites this deficiency as the impetus to develop another method for stochastic simulation.

> The simulation proceeds only forward in time, there is no way to account in advance for evidence known to have occurred until variables corresponding to these observations are brought into play and get sampled. If they match the observed data, the run is counted; otherwise, it must be discarded. The result is that the scheme requires an excessive number of simulation runs. In cases comprising large numbers of observations (e.g., 20), all but a small fraction of the simulations may be discarded, especially when a rare combination of data occurs. ( [Pearl 87])

Pearl's solution uses *local* numerical computation followed by logical sampling:

> A more desirable way of accounting for the evidence would be to permanently *clamp* the evidence variables to the values observed, then conduct stochastic simulation on the clamped network.
>
> The first step involves computing, for some variable X, its conditional distribution, given the states of all its neighbor variables. The second phase involves sampling the distribution computed in step 1 and instantiating X to the state selected by the sampling. The cycle then repeats itself by sequentially scanning through all the variables in the system. ( [Pearl 87])

The nodes are processed in arbitrary order, and can be initially set to any arbitrary value. Pearl describes an algorithm that initially sets all unknown variables to TRUE.

## 2.2. Local Intransigence

Although the simulation technique described by Pearl seems to work well with Bayesian belief networks that have nodes that are not highly dependent on one other, a problem arises when simulating graphs that contain nodes that are strongly dependent on one another. This problem can be illustrated by an example. Given the belief network shown in Figure 2-1, our task is to compute the posterior probability of all the nodes given that node E is true.[1]

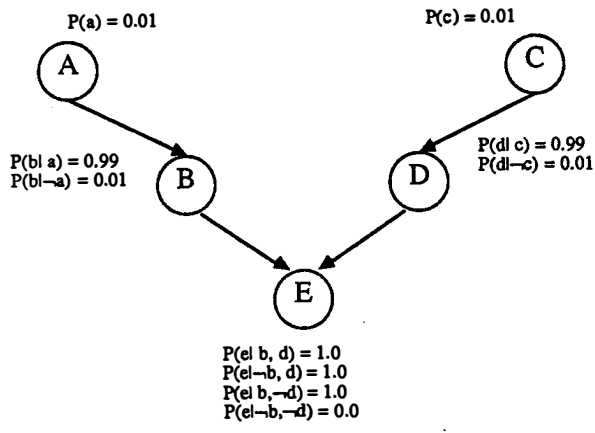

Figure 2-1: A five-node belief network that exhibits local intransigence.

Nodes A and C both have prior probabilities of TRUE of 0.01. These nodes in turn are closely linked to nodes B and D. Node B is TRUE with probability 0.99 if node A is TRUE, and Node B is FALSE with probability 0.99 if node A is FALSE. Nodes C and D are similarly dependent. Finally, node E is TRUE if *either* node B or D is TRUE, and FALSE only if both B and D are both FALSE.

---

[1]Uppercase letters represent variables, and lowercase letters represent instantiated variables. For example, A might represent the variable *brain tumor*, with a representing *brain tumor* = TRUE, and ¬a representing *brain tumor* = FALSE.

Using Pearl's algorithm, we initially set all the nodes to TRUE and proceed with stochastic simulation for each variable by local sampling of its neighbors. Because of the symmetry of the graph, the actual probability of TRUE for node A is equal to that of node C. Similarly, the actual probability of nodes B and D are equal.

Consider that, after a number of simulations, a state is reached where nodes B and E are TRUE, and nodes C and D are FALSE. Stochastic simulation at node B will give a probability of B remaining true of[2]

If node A = a:
$$P(B|W_B) = \frac{P(b|a)P(e|b,\neg d)}{P(b|a)P(e|b,\neg d) + P(\neg b|a)P(e|\neg b,\neg d)} = \frac{0.99*1}{0.99*1 + 0.01*0} = 1$$

If node A = ¬a:
$$P(B|W_B) = \frac{P(b|\neg a)P(e|b,\neg d)}{P(b|\neg a)P(e|b,\neg d) + P(\neg b|\neg a)P(e|\neg b,\neg d)} = \frac{0.01*1}{0.01*1 + 0.99*0} = 1$$

That is, on average, the probability of B remaining TRUE is 1, independent of the value of node A. When we stochastically simulate node C, we find the probability of it switching from FALSE to TRUE to be

$$P(C|W_C) = \frac{P(c)P(\neg d|c)}{P(c)P(\neg d|c) + P(\neg c)P(\neg d|\neg c)} = \frac{0.01*0.01}{0.01*0.01 + .99*.99} \approx 0.0001$$

That is, the probability of node C switching from FALSE to TRUE is 0.0001. Stochastic simulation of node D (assuming that C stays FALSE) shows that the probability of it becoming TRUE to be

$$P(D|W_D) = \frac{P(d|\neg c)P(e|d,b)}{P(d|\neg c)P(e|d,b) + P(\neg d|\neg c)P(e|\neg d,b)} = \frac{0.01*1}{0.01*1 + 0.99*1} = 0.01$$

That is, node D will stay FALSE 99 times out of 100, and turn TRUE 1 time out of 100. We can see that nodes B, C, and D have become relatively fixed, with a probability of changing of 1 in 100 or less, for any given simulation. When node D does becomes TRUE, simulation at node A gives

$$P(A|W_A) = \frac{P(a)P(b|a)}{P(a)P(b|a) + P(\neg a)P(b|\neg a)} = \frac{0.01*0.99}{0.01*0.99 + 0.99*0.01} = 0.5$$

and simulation at node B gives

If node A = a:
$$P(B|W_B) = \frac{P(b|a)P(e|b,d)}{P(b|a)P(e|b,d) + P(\neg b|a)P(e|\neg b,d)} = \frac{0.99*1}{0.99*1 + 0.01*1} = 0.99$$

If node A = ¬a:
$$P(B|W_B) = \frac{P(b|\neg a)P(e|b,d)}{P(b|\neg a)P(e|b,d) + P(\neg b|\neg a)P(e|\neg b,d)} = \frac{0.01*1}{0.01*1 + 0.99*1} = 0.01$$

Since node A has a 50:50 probability of being TRUE, node B has a 50:50 chance of being TRUE (0.5*0.99 + 0.5*0.01). Assuming that A and B do become FALSE in this iteration, we can see from the symmetry of the graph that D becomes fixed at TRUE, and A and B are now relatively fixed at the value FALSE. Due to the strong dependency between nodes, local areas in the graph tend to become *fixed* through many iterations rather than to change randomly according to their true probabilities.

The central reason for the development of this intransigence to change is illustrated by the simple network of two nodes in Figure 2-2. We can see that given the apriori probability of node A as 0.5, and given the symmetry of the probability matrix linking the two nodes, the posterior probability of both node A and B should be 0.5. However, once node A has been set to TRUE, the probability that node B is TRUE is 0.999, and the probability of node A staying TRUE with the next iteration (assuming that B has been set to TRUE) is 0.999. Both nodes will remain TRUE for an average of about 500 iterations. Any sampling of less than 500 iterations might well give the average probability of node A and B being TRUE as 1. Once

---

[2] $W_A$ represents the state of all other variables in the network except A. Thus $P(a|W_A)$ denotes the probability of A = TRUE, given the state of all the other variables in the network.

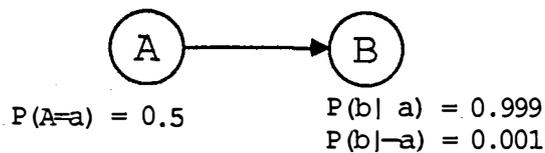

P(A=a) = 0.5   P(b| a) = 0.999
               P(b|¬a) = 0.001

Figure 2-2: A two-node belief network that exhibits local intransigence

either node A or B becomes FALSE (on the average, this will occur once in 500 iterations), the other node becomes FALSE with probability 0.999, and the two nodes again become relatively fixed. Instead of quick stochastic convergence to the true posterior probability, the calculated probability from the stochastic simulation *oscillates* toward the true value, with convergence at infinity.

Contrary to Pearl's claim that "dozens of runs are necessary for achieving reasonable levels of accuracy" ( [Pearl 87]), the number of runs that will be needed is strongly dependent on the probabilistic dependence between nodes in a causal graph. For the two-node example given here, convergence around the true probability of 0.5 occurs 500 times slower than it would if the simulation was done truly at random (for example, using Henrion's algorithm in this particular case).

In the limit, when there is a node pair that is deterministically dependent, Pearl's simulation algorithm fails completely. Once the deterministically dependent node pair is fixed at a certain value, each node will fix the other at that value for all subsequent iterations.

The problem that occurs with deterministically dependent nodes can be overcome by considering a group of deterministic nodes as one node for the purposes of simulation. The influences on any of the nodes in the group can be treated as an influence on the group as a whole, and the value of all the nodes in the group are determined with one simulation. Although this overcomes the limiting case of deterministically dependent nodes, this does not obviate the problem, in the worst case, that occurs with nodes that are strongly, but not categorically, dependent.

### 2.3. Number of Simulations Required in a Two-node Network

For a two-node network, the number of simulations required for convergence of a calculated probability to within a specified tolerance of the true probability increases exponentially as a function of the dependency between two nodes, reaching infinity at the limit where nodes are deterministically dependent. This relationship is seen in Figure 2-3.

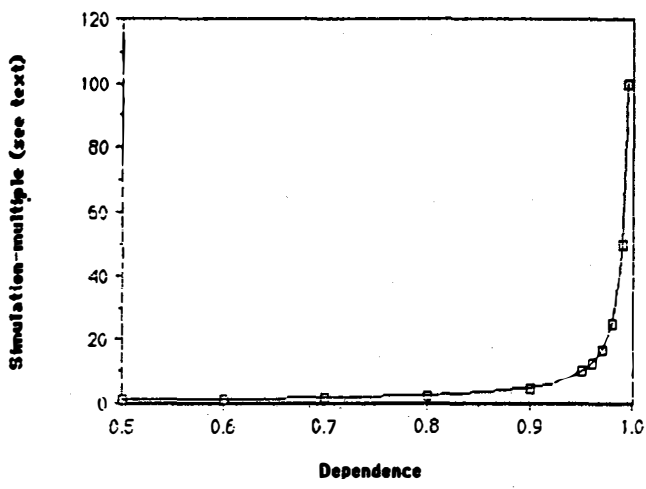

Figure 2-3: The Simulation-multiple as a
function of strength of dependence between two nodes.

The *Simulation-multiple* (SM) on the vertical axis is the multiple of the simulations that would be needed if the simulations were truly random. In other words, if 100 *random* simulations would be needed to achieve the necessary level of accuracy, and the Simulation-multiple is 40, then, using Pearl's algorithm, 4000 (100 times 40) simulations would be needed to achieve this same level of accuracy.

The horizontal axis is the *dependence* (D) between two nodes. The dependence is defined by the equation

$D = \sum \text{Min}[p_i, 1-p_i]$ over all the $p_i$ probabilities that link the two nodes.

A dependence value of 1 is the equivalent of random simulation. The dependence is highly influenced by the weakest probability that links the two nodes; for any two-node network, the Simulation-multiple is equal to the inverse of the dependence (1/D).

The nodes shown in Figure 2-2 have a dependence value of 0.002 (0.001 plus 0.001), and would require 500 times the number of simulations that would be needed if the simulations were done totally at random. In Figure 2-4, the weakest probability in the link matrix is $P(b|\neg a) = 0.5$. In this case, D = 0.501 (0.5 plus 0.001), and SM = 2.

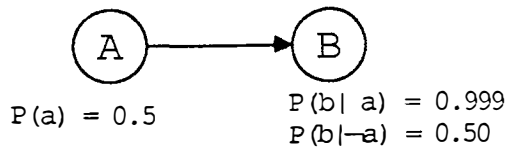

$P(a) = 0.5$  $P(b| a) = 0.999$
$P(b|\neg a) = 0.50$

Figure 2-4: A well-behaved two-node belief network.

### 2.4. Extension of Results to a Multiply-Connected Network

In an analogous way, these results can be extended to a more general belief network. The worst case local intransigence for any node can be calculated by the formula

$$D = \sum \prod \text{Min}[p_i, 1-p_i]$$ over all the $p_i$ probabilities in the Markov blanket for that node[3].

The strength of this dependence, and thus the severity of local intransigence, can thus be a product of the dependences in multiple link matrices in a given Markov blanket.

## 3. Modification of the Bayesian Belief Network

The use of Pearl's algorithm for the network depicted in Figure 2-1 results in *fixation* of node values for 100 simulations on average. That is, the convergence to the true value occurs 100 times more slowly than would be expected if the simulation was truly random. The use of Henrion's logic sampling algorithm, on the other hand, results in one usable sample for approximately every 25 simulations. The *weighted sampling* technique described by Henrion [Henrion 86] would yield many more usable samples; but in general it is, in the worst case, still exponential in the number of observed variables. For n observed binary variables, 1 out of $2^n$ simulations would be usable.

Methods to manipulate influence diagrams have been discussed extensively by Shachter [Shachter 86, Shachter 87] in the context of influence diagram solution. One of the major problems in finding an exact solution to an influence diagram is that the time complexity to solution may be $O(2^N)$, where N is the number of nodes in the graph. By applying the techniques described in [Shachter 87] to local areas in the belief network it may be possible to transform the belief network into one that will generate stochastic sample instances more efficiently. The techniques of graph pruning, arc reversal, and node reduction, in the context of stochastic simulation, are discussed in the following sections.

### 3.1. Graph Pruning

For a given problem it may be possible to prune the graph to decrease the number of nodes that need to be simulated for that particular problem. If J is the set of nodes for which values are known, and K is the set of nodes for which probabilities are sought, then any node x in the belief network can be removed without affecting the solution, if a directed path cannot be drawn from node x to any of the nodes in the set J ∪ K (see [Shachter 87] for formal proof). For example, Figure 3-1a shows an example where no pruning can be done, and the entire graph must be used in stochastic simulation. In Figure 3-1b, the nodes in set A can be pruned away as no directed path exists from any of the nodes in set A to any of the nodes in J or K. In Figure 3-1c, both the nodes in set A and B can be pruned, and stochastic simulation need only be done on the remaining graph.

### 3.2. Arc Reversal

For any given belief network, it is possible to reverse arcs that are pointing into known conditioning nodes, until the only arcs pointing into known nodes are arcs from other known nodes. The advantage of such a network is that the nodes with known values can now be fixed at those values, and Henrion's algorithm can be applied. Because each simulated instance will have the appropriate values for nodes with known values, each simulation will produce an appropriate sample instance.

---

[3] A Markov blanket is the union of three type of neighbors: direct parents, direct successors and all direct parents of the latter (see [Pearl 87] for a more complete discussion).

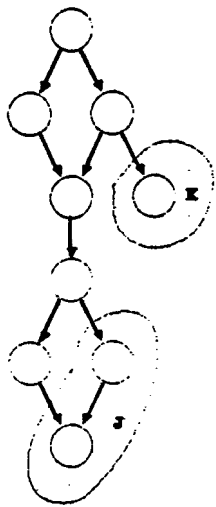 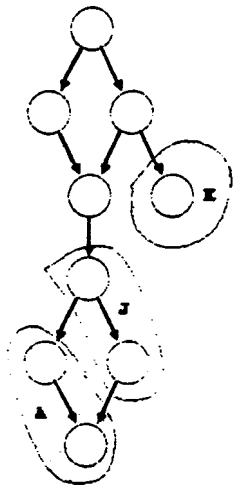 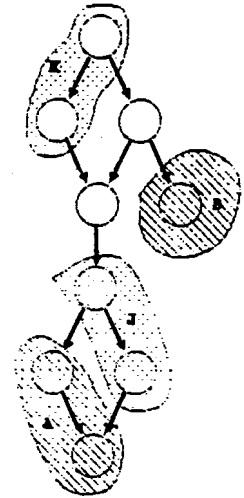

Figure 3-1a          Figure 3-1b          Figure 3-1c

Figure 3-1: Examples of graph pruning.

The process of arc-reversal between two nodes A and B involves the calculation of the new conditional dependencies between node B and the immediate predecessors (if any) of node A, as well as the new conditional dependencies between node A and the immediate predecessors of node B. These dependencies are represented by new arcs where necessary (see [Shachter 87] for details). Since the complexity of arc-reversal is dependent on the number of predecessors of the nodes at each end of the arc, the applicability of this methodology will depend on the number of total predecessors (*weak predecessors* in [Shachter 87]) of the two nodes at the ends of the arc being reversed. For this reason, certain graphs will be more amenable to this technique than others, and for any given graph, it may be beneficial to reverse some arcs and not others.

In addition, the optimal strategy for a given problem will depend on the number of simulations that are necessary to achieve the specified degree of accuracy. The use of arc reversal is analogous to the economic concept of fixed costs (the computational cost of arc-reversal) that will reduce future variable costs (the decrease in the computational cost needed for each usable simulation). If only a limited number of simulations are needed, the optimal strategy might be logic sampling of the original network; if the number of samples needed is greater than a certain threshold, the use of arc reversal will become increasingly beneficial. Figure 3-2a shows a network with four known conditioning nodes (set J), and one node for which we wish to determine the probability (set K). If strong dependencies exist between nodes, the use of both Pearl's and Henrion's algorithms may be inefficient. Even the use of Henrion's weighted sampling scheme, where probabilities between node values are set to 0.5 and appropriate adjustments are made to the weight of the node value, will result in only 1 usable sample for every 16 simulations ($0.5^4$). This is because the simulation begins with the instantiation of nodes A and B first. When the nodes in the set J are instantiated, only 1 sample out of 16 will have been instantiated to the appropriate values necessary for the simulation to be included as a positive instance.

By the use of arc reversal, the network can be modified to that seen in Figure 3-2b. Nodes A and B can now be pruned (as described in the previous section), and the network in Figure 3-2c is the result. From this network, the known nodes can be set to their known values, and each simulation will result in a usable instance.

### 3.3. Node Reduction
By removing nodes that have arcs with highly dependent values through the method of node reduction [Shachter 87], it may be possible to reduce the dependency between nodes to decrease the local intransigence that can be seen in the use of Pearl's algorithm. For example, given the three nodes in Figure 3-3a, the use of Pearl's algorithm will result in convergence that is 500 times slower than a completely random sampling. By the process of node reduction, node B can be removed, and the conditional probabilities propagated into node C. The resulting conditional probabilities for node C are shown in Figure 3-3b. The strong dependence that existed between nodes B and C, which would have resulted in local intransigence, has now been eliminated, resulting in a much more random sampling when using Pearl's algorithm.

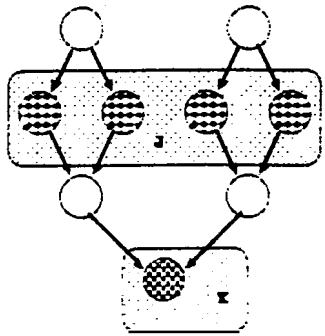 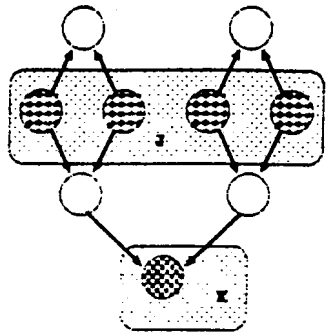 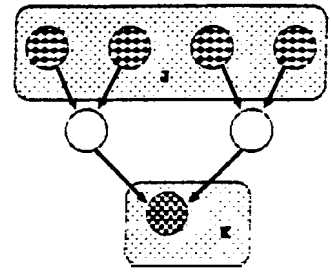

Figure 3-2a    Figure 3-2b    Figure 3-2c

Figure 3-2: An example of the use of arc reversal to improve Henrion's algorithm for stochastic simulation in a belief network.

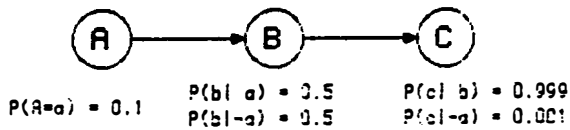 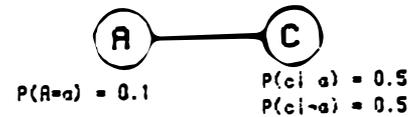

Figure 3-3a    Figure 3-3b

Figure 3-3: The use of node reduction to improve Pearl's algorithm for the stochastic simulation of a Bayesian belief network.

## 4. Summary

We have examined two previously described methods for performing stochastic simulation. We found that a method developed by Pearl had previously undescribed problems, and in certain cases, required a much larger number of simulations than expected to obtain convergence of the calculated probability. A method described by Henrion has similar shortcomings in that numerous simulations are needed to generate one appropriate sample instance. In this paper we explore the use of network-manipulation techniques to modify the Bayesian belief network into a form that will more efficiently generate appropriate cases with stochastic simulation. Specifically, we discuss three techniques: Graph pruning, arc-reversal, and node reduction. Graph pruning has wide applicability to all methods of solution. Arc-reversal results in significant improvement, in some cases, using Henrion's technique. The use of node reduction can in some cases yield significant benefits for Pearl's algorithm. The delineation of the trade-offs and criteria for the applicability of each method is the focus of current research.

## Acknowledgements

This work has been supported by grant LM-07033 from the National Library of Medicine. Computer facilities were provided by the SUMEX-AIM resource under NIH grant RR-00785.